\documentclass[11pt]{article}

\usepackage[final]{acl}
\usepackage{times}
\usepackage{latexsym}
\usepackage{enumitem}
\definecolor{mypink}{rgb}{.99,.91,.95}
\definecolor{mygray}{rgb}{0.9, 0.9, 0.9}
\definecolor{linkblue}{rgb}{0.2,0.30,0.568}

\usepackage{hyperref}
\usepackage{url}
\usepackage{amssymb}
\usepackage{graphicx}
\usepackage{booktabs}
\usepackage{algorithm} 
\usepackage{algpseudocode}
\usepackage{mathrsfs}
\usepackage{multirow}
\usepackage{color}
\usepackage{colortbl}
\usepackage{arydshln}
\usepackage{array}
\usepackage{bm}
\usepackage{rotating}
\usepackage{adjustbox}
\usepackage{color}

\usepackage{textcomp}
\usepackage{mdframed}
\usepackage{amsmath}
\usepackage{CJKutf8}
\usepackage{algorithm} 
\usepackage{balance}
\usepackage{algpseudocode}

\usepackage{pifont}
\usepackage{graphicx}   
\usepackage{tikz}       
\usepackage[edges]{forest} 
\usepackage{graphicx}
\usepackage{multirow}
\usepackage{multicol}
\usepackage{tabularx}
\usepackage{booktabs}
\usepackage{adjustbox}

\newcommand{\cmark}{\ding{51}}

\definecolor{hidden-red}{RGB}{205, 44, 36}
\definecolor{hidden-blue}{RGB}{194,232,247}
\definecolor{hidden-orange}{RGB}{243,202,120}
\definecolor{hidden-green}{RGB}{34,139,34}
\definecolor{hidden-pink}{RGB}{255,245,247}
\definecolor{hidden-black}{RGB}{20,68,106}

\definecolor{hidden-red}{RGB}{205, 44, 36}
\definecolor{hidden-blue}{RGB}{194,232,247}
\definecolor{hidden-orange}{RGB}{243,202,120}
\definecolor{hidden-green}{RGB}{34,139,34}
\definecolor{hidden-pink}{RGB}{255,245,247}
\definecolor{hidden-black}{RGB}{20,68,106}

\definecolor{myGreen}{RGB}{127,210,85}
\definecolor{myOrange}{RGB}{242,154,66}
\definecolor{myYellow}{RGB}{247,223,65}
\definecolor{myRed}{RGB}{232,80,43}
\definecolor{myViolet}{RGB}{162,57,102}
\definecolor{myBlue}{HTML}{4686f3}
\definecolor{myYellowv2}{HTML}{E6C802}
\definecolor{myOrangev2}{HTML}{ED8E55}
\definecolor{MyGreenv2}{HTML}{009B55}
\definecolor{MyRedv2}{HTML}{c22f2f}
\definecolor{DarkRed}{RGB}{130,25,0}
\definecolor{PurpleRed}{RGB}{204,0,102}
\definecolor{DarkGreen}{RGB}{30,130,30}
\definecolor{DarkBlue}{RGB}{0,0,250}
\definecolor{DarkYellow}{RGB}{255,128,0}

\usepackage[T1]{fontenc}

\usepackage{microtype}

\usepackage{inconsolata}

\usepackage{graphicx}

\title{Test-Time Scaling in Multimodal Foundation Models: A Comprehensive Survey of Generation and Reasoning}

\author{
Cong Wan\textsuperscript{1},
Ying He\textsuperscript{1},
Zhongzhan Huang\textsuperscript{1},
Hefeng Wu\textsuperscript{1}\thanks{Corresponding author is Hefeng Wu.}
\\
\textsuperscript{1}Sun Yat-sen University
\\
\texttt{\{wanc,heying63\}@mail2.sysu.edu.cn}
\\
\texttt{zhongzhanhuang@foxmail.com}
\quad
\texttt{wuhefeng@gmail.com}
}

\begin{document}

\maketitle

\begin{abstract}

Test-time Scaling (TTS) has emerged as a pivotal research direction for enhancing model performance by dynamically allocating computational resources during inference. Recent advancements have adapted this paradigm to Multimodal Foundation Models (MFMs), unlocking their potential in multimodal reasoning and generation.
Despite rapid progress, the field lacks a systematic survey and unified theoretical framework to delineate the developmental landscape of multimodal TTS.
To bridge this gap, we present the first comprehensive review of TTS research for MFMs, proposing a unified taxonomic framework that categorizes existing methodologies into three distinct strategies: sampling-based, feedback-based, and search-based approaches.
We further summarize representative applications and benchmarks commonly utilized to evaluate multimodal TTS capabilities in generation and reasoning tasks.
Finally, this survey discusses open challenges and outlines future research directions, providing a systematic roadmap for subsequent studies in this rapidly evolving field.

\end{abstract}
\label{sec:}%

\section{Introduction}\label{sec:intro}%

Foundation Models have revolutionized the landscape of artificial intelligence, particularly in generation and reasoning tasks~\citep{bommasani2022opportunitiesrisksfoundationmodels}. As the cornerstone of this paradigm, Large Language Models (LLMs) have achieved remarkable advancements \citep{brown2020language,achiam2023gpt,HuangLZWL25acl}, demonstrating exceptional emergent abilities in complex reasoning~\citep{wei2022chain}.
This success is primarily driven by scaling model parameters, data volume, and computational resources during pre-training to enhance reasoning capabilities. 
This training-phase scaling behavior is termed scaling laws~\citep {kaplan2020scaling}, providing theoretical guidance for large-scale model training.

\begin{figure*}[t!]
    \centering
\includegraphics[width=\textwidth]{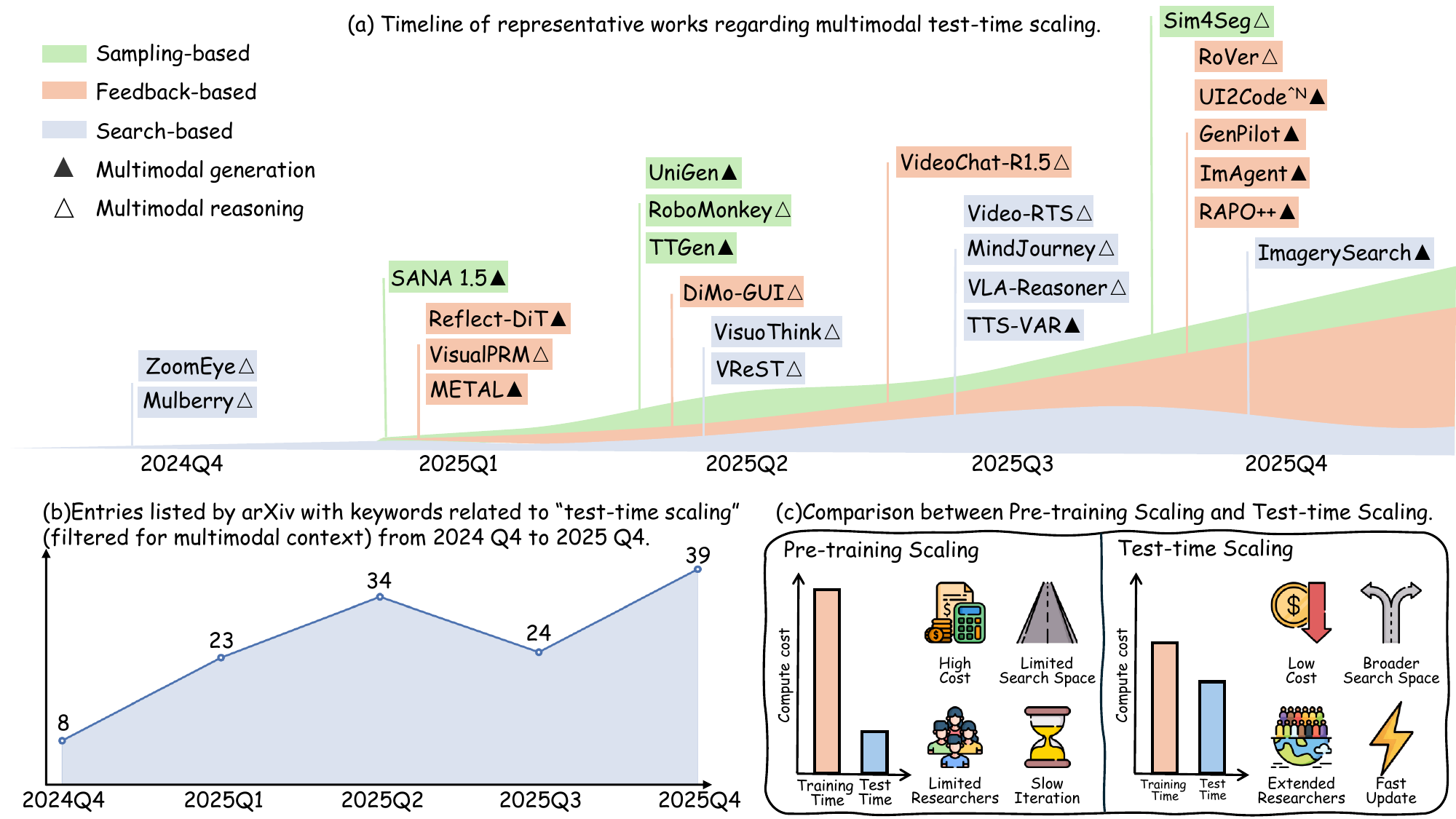}
    \caption{Recent trends in multimodal test-time scaling regarding historical evolution, publication growth, and the paradigm shift from pre-training. Q denotes quarter.}
    \label{fig:timeline}
\end{figure*}

However, recent studies suggest that relying solely on increased training data and compute is yielding diminishing marginal returns~\citep {diaz2024scaling}.
Consequently, research focus has shifted toward unlocking the latent potential of strong foundation models during the inference phase. 
In this context, Test-time Scaling (TTS)~\citep {snell2024scaling,wu2024inference} has emerged as a promising paradigm complementary to pre-training scaling.
Unlike pre-training scaling, TTS dynamically allocates computational resources during inference to exploit model capabilities without additional parameter updates, as illustrated in Fig.~\ref{fig:timeline}(c).
Specifically, prior works have systematically explored TTS strategies, such as search~\citep{xie2024monte}, sampling~\citep {chow2024inference}, and verification~\citep {hosseini2024v}, to enhance LLM reasoning.
These efforts not only demonstrate the efficacy of TTS in LLMs but also inspire its extension to Multimodal Foundation Models (MFMs)~\citep {li2024multimodal}.

Inspired by the success of LLMs, the research community is rapidly shifting focus toward MFMs integrating diverse modalities such as vision and language.
These models refer to general-purpose foundations capable of processing multimodal data, encompassing cross-modal understanding architectures based on Multimodal Large Language Models (MLLMs) and vision-language generation frameworks typified by Diffusion Models.
Through cross-modal fusion and alignment, MFMs are pivotal for the realization of Artificial General Intelligence (AGI).
However, similar to LLMs, further performance improvements of MFMs are also constrained by scaling laws. 
Leveraging the efficacy of TTS in LLMs, researchers are adapting similar strategies to MFMs to further unlock latent potential during inference, obviating the need for additional training or parameter expansion.
As illustrated in Fig.~\ref{fig:timeline}, this paradigm shift has triggered a surge in research interest and a rapidly evolving landscape of methodologies.

Current research on TTS for MFMs centers on two domains: multimodal generation~\citep {ma2025inference,xie2025sana} and multimodal reasoning~\citep {zhu2025internvl3}.
Within these tasks, strategies such as majority voting~\citep {byun2025test}, tree search~\citep {liu2025video}, and reward models~\citep {qiao2025ttgen} are being actively adapted from language models to multimodal tasks.

Despite the surge in interest, a unified taxonomy that synthesizes TTS advancements within multimodal scenarios remains absent.
While prior surveys predominantly focus on TTS for LLMs~\citep {zhang2025survey,ji2025surveytesttimecomputeintuitive}, to the best of our knowledge, this work represents the first comprehensive review dedicated to MFMs.
By systematically organizing recent advancements, we aim to provide a clear roadmap and reference for future research.

The structure of this work is organized as follows. 
We begin by introducing MFM techniques and TTS background in Sec.~\ref{sec:Background}. 
Next, in Sec.~\ref{sec:Multimodal Test-time Scaling}, we establish a unified taxonomy, systematically classifying approaches into sampling-, feedback-, and search-based paradigms. 
Sec.~\ref{sec:Applications and Benchmarks} then details specific applications in multimodal generation and reasoning, with relevant benchmarks provided in the Appendix~\ref{sec:Benchmarks}.
Finally, we offer a comprehensive discussion in Sec.~\ref{sec:Challenges and Future Directions}, highlighting open challenges and future directions for research.

Our main contributions are threefold:

\textbullet~\textbf{First Systematic Survey.} We present the first systematic survey dedicated to TTS in MFMs, bridging a critical gap in the current literature.

\textbullet~\textbf{Unified Taxonomy.} We propose a structured taxonomy categorizing existing methods, clarifying their mechanisms and applicability.

\textbullet~\textbf{Future Roadmap.} We analyze relevant benchmarks and highlight open challenges, offering a strategic guide for future research.

\tikzstyle{my-box}=[
    rectangle,
    draw=hidden-black,
    rounded corners,
    text opacity=1,
    minimum height=1.5em,
    minimum width=5em,
    inner sep=2pt,
    align=center,
    fill opacity=.5,
]
\tikzstyle{leaf}=[
    my-box, 
    minimum height=1.5em,
    fill=hidden-blue!90, 
    text=black,
    align=left,
    font=\normalsize,
    inner xsep=2pt,
    inner ysep=4pt,
]

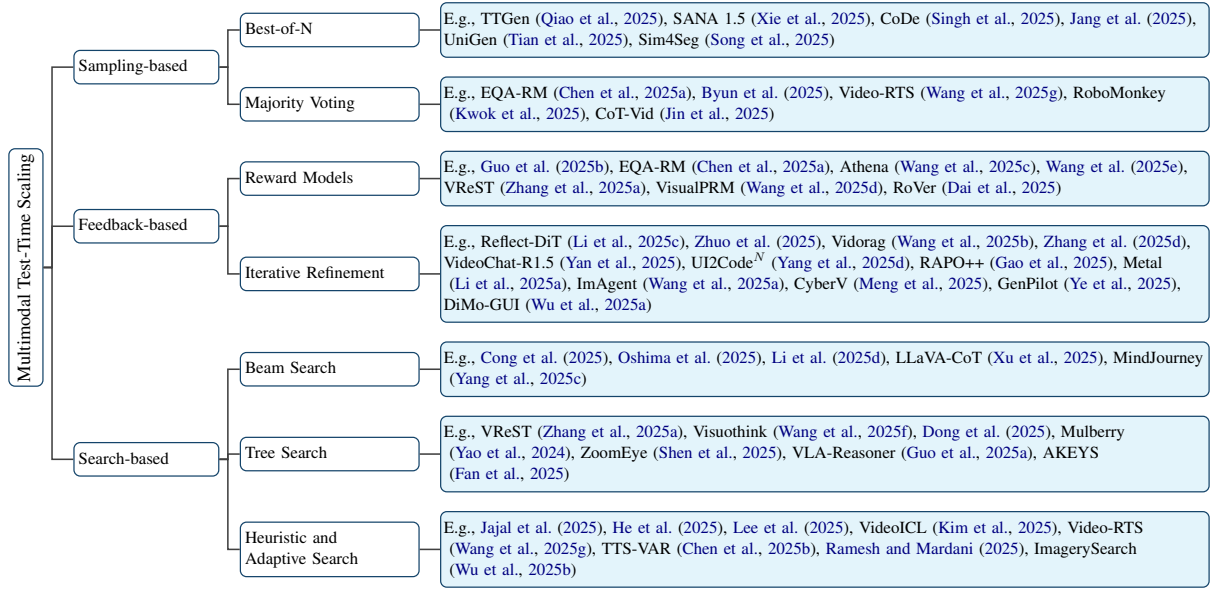
\begin{figure*}[t]
    \centering
    \resizebox{\textwidth}{!}{
        \begin{forest}
            forked edges,
            for tree={
                grow=east,
                reversed=true,
                anchor=base west,
                parent anchor=east,
                child anchor=west,
                base=left,
                font=\large,
                rectangle,
                draw=hidden-black,
                rounded corners,
                align=left,
                minimum width=4em,
                edge+={darkgray, line width=1pt},
                s sep=12pt,
                inner xsep=3pt,
                inner ysep=4pt,
                line width=0.8pt,
                ver/.style={rotate=90, child anchor=north, parent anchor=south, anchor=center},
            },
            where level=1{text width=8em,font=\normalsize,}{},
            where level=2{text width=10em,font=\normalsize,}{},
            [
                Multimodal Test-Time Scaling, ver
                [ 
                    Sampling-based
                    [
                        Best-of-N
                        [
                            {E.g.,} 
                            TTGen~\citep{qiao2025ttgen}{,}
                            SANA 1.5~\citep{xie2025sana}{,}
                            CoDe~\citep{singh2025code}{,}
                            \citet{jang2025verifier}{,}\\
                            UniGen~\citep{tian2025unigen}{,}
                            Sim4Seg~\citep{song2025sim4seg}
                            , leaf, text width=45em
                        ]
                    ]
                    [
                        Majority Voting
                        [
                            {E.g.,} 
                            EQA-RM~\citep{chen2025eqa}{,}
                            \citet{byun2025test}{,}
                            Video-RTS~\citep{wang2025video}{,}
                            RoboMonkey\\~\citep{kwok2025robomonkey}{,}
                            CoT-Vid~\citep{jin2025cot}
                            , leaf, text width=45em
                        ]
                    ]
                ]
                [
                    Feedback-based
                    [
                        Reward Models
                        [
                            {E.g.,} 
                            \citet{guo2025can}{,}
                            EQA-RM~\citep{chen2025eqa}{,}
                            Athena~\citep{wang2025athena}{,}
                             \citet{wang2024scaling}{,}\\
                             VReST~\citep{zhang2025vrest}{,}
                             VisualPRM~\citep{wang2025visualprm}{,}
                             RoVer~\citep{dai2025rover}
                            , leaf, text width=45em
                        ]
                    ]
                    [
                        Iterative Refinement
                        [
                            {E.g.,} 
                            Reflect-DiT~\citep{li2025reflect}{,}
                            \citet{zhuo2025reflection}{,}
                            Vidorag~\citep{wang2025vidorag}{,}
                            \citet{zhang2025generative}{,}\\
                            VideoChat-R1.5~\citep{yan2025videochat}{,}
                            UI2Code$^N$~\citep{yang2025ui2code}{,}
                            RAPO++~\citep{gao2025rapo++}{,}
                            Metal\\~\citep{li2025metal}{,}
                            ImAgent~\citep{wang2025imagent}{,}
                            CyberV~\citep{meng2025cyberv}{,}
                             GenPilot~\citep{ye2025genpilot}{,}\\
                             DiMo-GUI~\citep{wu2025dimo}
                            , leaf, text width=45em
                        ]
                    ]
                ]
                [
                    Search-based
                    [
                        Beam Search
                        [
                            {E.g.,} 
                            \citet{cong2025can}{,}
                            \citet{oshima2025inference}{,}
                            \citet{li2025dynamic}{,}
                            LLaVA-CoT~\citep{xu2025llava}{,}
                            MindJourney\\~\citep{yang2025mindjourney}
                            , leaf, text width=45em
                        ]
                    ]
                    [
                        Tree Search
                        [
                            {E.g.,} 
                            VReST~\citep{zhang2025vrest}{,}
                            Visuothink~\citep{wang2025visuothink}{,}
                             \citet{dong2024progressive}{,}
                            Mulberry\\~\citep{yao2024mulberry}{,}
                            ZoomEye~\citep{shen2025zoomeye}{,}
                            VLA-Reasoner~\citep{guo2025vla}{,}
                            AKEYS\\~\citep{fan2025agentic}
                            , leaf, text width=45em
                        ]
                    ]
                    [
                        Heuristic and \\Adaptive Search
                        [
                            {E.g.,} 
                            \citet{jajal2025inference}{,}
                             \citet{he2025scaling}{,}
                            \citet{lee2025adaptive}{,}
                            VideoICL~\citep{kim2025videoicl}{,}
                            Video-RTS\\~\citep{wang2025video}{,}
                            TTS-VAR~\citep{chen2025tts}{,}
                            \citet{ramesh2025test}{,}
                            ImagerySearch\\~\citep{wu2025imagerysearch}
                            , leaf, text width=45em  
                        ]
                    ]
                ]
            ]
        \end{forest}
    }
    \caption{Taxonomy of multimodal test-time scaling methods. }
    \label{fig:task_taxonomy}
\end{figure*}

\vspace{5pt}

\section{Background}\label{sec:Background}

\subsection{Preliminaries of MFMs}
This section outlines the preliminaries of MFMs. We specifically focus on MLLMs and Diffusion Models, as they serve as the primary foundations where test-time scaling strategies are currently explored and applied.

\paragraph{Multimodal Large Language Models (MLLMs)}
MLLMs, exemplified by GPT-4V~\citep{achiam2023gpt} and Gemini 2.0~\citep{team2023gemini}, have converged on a unified architectural paradigm.
In understanding-centric MLLMs~\citep{li2023blip,liu2023visual}, visual, audio, or video signals are typically encoded into continuous embeddings or discrete tokens and mapped to the decoder's input space, enabling autoregressive processing analogous to text generation.
This unification provides the architectural foundation for scaling test-time computation via Chain-of-Thought (CoT) reasoning or multi-step verification.

Building on this, MLLMs with multimodal generation capabilities extend this paradigm, primarily by either discretizing multimodal outputs for direct autoregressive token generation~\citep{zhan2024anygpt} or routing multimodal embeddings to specific decoders~\citep{wu2024next}.
Fundamentally, these architectures reframe multimodal generation as a sequential decision-making or feature planning problem.  Consequently, generation transitions from a unidirectional output to an optimizable inference process amenable to test-time search, verification, or iterative refinement, thereby catalyzing research into TTS for MLLMs.

\paragraph{Diffusion Models}
Diffusion models~\citep{song2020score,ho2022classifier} have demonstrated superior performance in text-conditional generation, emerging as the dominant paradigm in visual generation.
Unlike single-step generative models, diffusion models synthesize data via an iterative denoising process. This iterative nature facilitates a flexible trade-off between compute budget (e.g., sampling steps, candidate population) and generation fidelity, providing intrinsic support for TTS.

Fundamentally, diffusion models degrade the complex data distribution $p_{\mathrm{data}}(\mathbf{x})$ into Gaussian noise $\mathcal{N}(0, \sigma^2 I)$ through a forward diffusion process.
Given a clean sample $x_0$, the latent state at noise level $t$ is formulated as:
\begin{equation}
x_t = \sqrt{\bar{\alpha}_t}x_0 + \sqrt{1-\bar{\alpha}_t}\epsilon, \quad \epsilon \sim \mathcal{N}(0, \mathrm{I}).
\end{equation}
where $\bar{\alpha}_t = \prod_{t=1}^T \alpha_t$ represents a predefined noise schedule and $\epsilon$ denotes standard Gaussian noise.
To reverse this, a network $\epsilon_\theta(x_t, t)$ is trained to predict the added noise $\epsilon$, enabling the gradual reconstruction of the data distribution.

In multimodal conditional generation, the noise predictor extends to a conditional form $\epsilon_\theta(x_t, t, \phi)$, typically employing Classifier-Free Guidance (CFG)~\citep{ho2022classifier} to enhance the alignment between the output and the condition.
CFG modulates the final prediction via linear extrapolation between conditional and unconditional estimates:
\begin{equation}
\begin{split}
    \hat{\epsilon}_t &= \epsilon_{\theta}(x_t, t, \phi_{\text{none}}) \\
    &\;\; + \omega_g \big(\epsilon_{\theta}(x_t, t, \phi_{\text{cond}}) - \epsilon_{\theta}(x_t, t, \phi_{\text{none}})\big).
\end{split}
\end{equation}
where $\phi_{\text{cond}}$ and $\phi_{\text{none}}$ denote the conditional (e.g., text prompt) and null embeddings, respectively, and $\omega_g$ is the guidance scale.
This explicit controllability serves as a critical foundation for advanced test-time optimization, such as sampling-based selection or iterative refinement.

\subsection{The Necessity of Test-time Scaling for MFMs}

\begin{table*}[t!]
\centering
\resizebox{\textwidth}{!}{
\begin{tabular}{ll l cc cc l} 
\toprule
\multirow{2}{*}{\textbf{Category}} & \multirow{2}{*}{\textbf{Method}} & \multirow{2}{*}{\textbf{Task}} & \multicolumn{2}{c}{\textbf{Domain}} & \multicolumn{2}{c}{\textbf{Guidance Signal}} & \multirow{2}{*}{\textbf{Approach Description}} \\
\cmidrule(lr){4-5} \cmidrule(lr){6-7} 
 & & & \textbf{Generation} & \textbf{Reasoning} & \textbf{Function} & \textbf{MLLM} & \\
\midrule

\multirow{6}{*}{\textbf{Best-of-N}} 
& TTGen~\citep{qiao2025ttgen} & Image Generation & \cmark &  & \cmark & & Use CLIP score to select the best latent during diffusion \\
& SANA 1.5~\citep{xie2025sana} & Image Generation & \cmark &  & & \cmark &Filters mismatches via VILA-Judge tournament and VLM scoring \\
& CoDe~\citep{singh2025code} & Image Generation & \cmark &  & \cmark & &Selects optimal diffusion outputs via block-based sampling \\
& UniGen~\citep{tian2025unigen} & Generation\&Reasoning& \cmark & \cmark & & \cmark &Combines generation and verification via BoN \\
&\citet{jang2025verifier}& Vision Language Action & &  \cmark & \cmark & &Selects actions via masked reference KL scoring\\  
& Sim4Seg~\citep{song2025sim4seg} & Medical Diagnosis & &  \cmark & \cmark & &Performs Best-of-N via joint semantic-visual scaling \\
\midrule
\multirow{5}{*}{\shortstack{\textbf{Majority} \\ \textbf{Voting}}} 
& EQA-RM \cite{chen2025eqa} & Embodied QA & & \cmark & \cmark &  &Generative reward model with majority voting \\
&  \citet{byun2025test} & Medical Diagnosis & & \cmark & \cmark &  & Aggregates visual descriptions via voting to reduce misdiagnosis \\
& Video-RTS~\citep{wang2025video} & Video Reasoning & & \cmark & \cmark &  & Enhance video reasoning via consistency voting \\
& RoboMonkey~\citep{kwok2025robomonkey} & Vision Language Action & & \cmark & \cmark &  &Select the optimal action via majority voting on perturbed samples \\
& CoT-Vid~\citep{jin2025cot} & Video Reasoning& & \cmark & \cmark & &Performs self-consistency via character-level clustering \\
\bottomrule
\end{tabular}
}
\caption{Summary of sampling-based methods. \textbf{Guidance Signal} distinguishes between \textbf{Function} (explicit scoring functions or statistical voting formulas) and \textbf{MLLM} (semantic judgment or aggregation).}
\label{tab:sampling_table}
\vspace{3pt}
\end{table*}

The necessity of TTS arises from its ability to overcome the prohibitive costs and static nature of traditional training paradigms. By dynamically allocating inference-time compute, TTS offers a cost-effective alternative to static parameter expansion and enables immediate adaptation to distribution shifts without requiring weight updates.
Crucially, single decoding paths often fail to capture the complex reasoning required in high-dimensional multimodal tasks. While TTS introduces mechanisms like search and verification to explore broader solution spaces, implementing these strategies in MFMs is fundamentally more challenging than in text-only LLMs.

Importantly, although TTS strategies in MFMs resemble those in LLMs, their multimodal instantiation is fundamentally more challenging. Unlike text-only models that allocate inference-time compute strictly over unimodal reasoning, MFMs must simultaneously scale compute across perceptual evidence, spatial grounding, and temporal context. Consequently, evaluating intermediate steps requires strict cross-modal faithfulness to visual and spatial relations, beyond mere textual consistency. Moreover, the inherent modality gap in multimodal generation often necessitates auxiliary VLMs or reward models. As a result, TTS for MFMs must scale not only reasoning depth, but also perception, grounding, and cross-modal verification.

\subsection{Scope and Formalization of TTS}

TTS can be formulated as selecting an inference procedure $\pi$ that queries a fixed model to maximize expected utility subject to a test-time compute budget:
\begin{equation}
\begin{aligned}
\pi^{*} = \arg\max_{\pi}\ & \mathbb{E}_{y \sim \pi(\cdot \mid x, \theta)} [U(x,y)] \\
\text{s.t.}\ & C(\pi, x) \le B,\quad \theta\ \text{fixed}.
\end{aligned}
\label{eq:tts_formulation}
\end{equation}
Here, $x$ denotes the input, $\theta$ denotes the model parameters fixed at test time, $y$ denotes the generated output, $U(x,y)$ denotes the task utility, and $C(\pi,x)$ denotes the test-time computational cost incurred by applying $\pi$ to $x$. This formulation highlights that TTS scales the inference procedure rather than the model parameters, while allowing the compute cost to vary across inputs.

To further clarify the scope of this survey, we distinguish three resources that may change at test time: compute, memory/state, and weights. In this work, TTS primarily refers to \emph{compute-centric} inference, where model parameters remain fixed and additional budget is allocated to operations such as sampling, search, verification, or iterative refinement. By contrast, test-time memory methods additionally modify dynamic state beyond standard one-pass inference, e.g., through retrieval stores, episodic memory, persistent caches, or expressive hidden states~\citep{suzgun2026dynamic}, whereas test-time training/adaptation updates model parameters through gradient-based or lightweight adaptation mechanisms~\citep{dalal2025one,liang2025comprehensive}. Some multimodal methods may involve both compute scaling and memory augmentation; in such cases, we classify them by the dominant scaling mechanism and treat memory as an auxiliary component. A compact comparison is provided in Appendix Table~\ref{tab:tts_scope}.

\section{Multimodal Test-time Scaling}\label{sec:Multimodal Test-time Scaling}
This section systematically reviews recent advancements in TTS for MFMs. As illustrated in Fig.~\ref{fig:task_taxonomy}, we categorize existing approaches into three distinct paradigms: sampling-based, feedback-based, and search-based methods, and further compare their applicability and trade-offs across multimodal tasks.

\subsection{Sampling-based Methods}\label{sec:Sampling-based Methods}

\begin{figure}[t!]
    \centering
\includegraphics[width=\columnwidth]{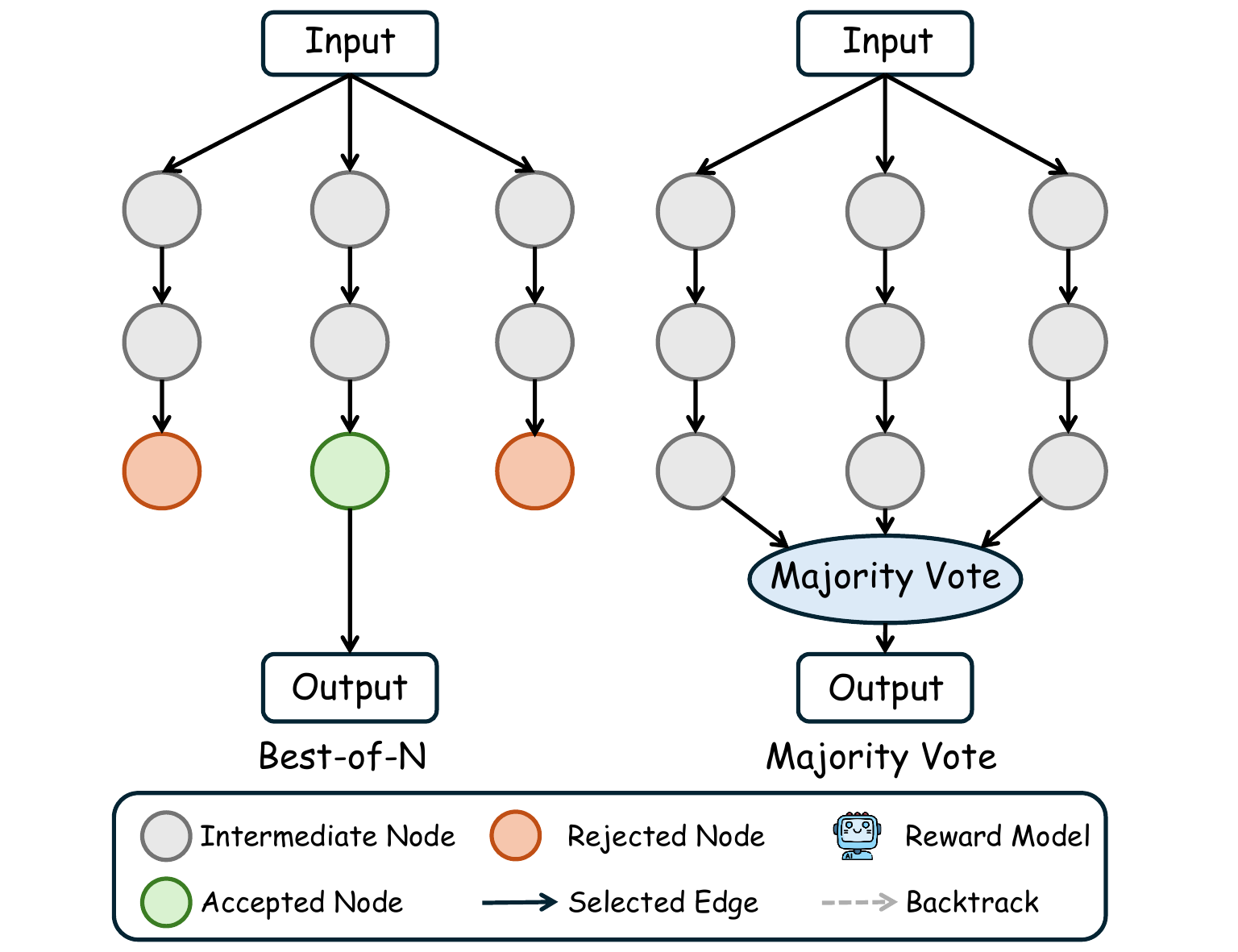}
    \caption{Illustration of sampling-based methods. }
    \label{fig:Sampling-zhixian Methods}
\end{figure}

Sampling-based methods explicitly scale test-time computation by generating multiple candidate solutions in parallel, employing aggregation or selection mechanisms to enhance output fidelity and diversity.
Compared to single-sample generation, such approaches explore a broader solution space, yielding superior performance in image generation, multimodal reasoning, and question-answering tasks.
Once candidates are generated, the critical step is selecting the final output.
As depicted in Fig.~\ref{fig:Sampling-zhixian Methods}, prevalent strategies primarily include Best-of-N (BoN) and Majority Voting.
A detailed taxonomy and summary of all surveyed methods are provided in Table~\ref{tab:sampling_table}.

\subsubsection{Best-of-N}\label{sec:Best-of-N}

\begin{table*}[t!]
\centering
\resizebox{\textwidth}{!}{
\begin{tabular}{ll l cc cc ccc l}
\toprule
\multirow{2}{*}{\textbf{Category}} & \multirow{2}{*}{\textbf{Method}} & \multirow{2}{*}{\textbf{Task}} & \multicolumn{2}{c}{\textbf{Domain}} & \multicolumn{2}{c}{\textbf{Feedback Scope}} & \multicolumn{3}{c}{\textbf{Feedback Form}} & \multirow{2}{*}{\textbf{Approach Description}} \\
\cmidrule(lr){4-5} \cmidrule(lr){6-7} \cmidrule(lr){8-10}
 & & & \textbf{Gen.} & \textbf{Reas.} & \textbf{Global} & \textbf{Step} & \textbf{Scalar} & \textbf{Text} & \textbf{Visual} & \\
\midrule

\multirow{7}{*}{\textbf{Reward Models}} 
&\citet{guo2025can} & Image Generation & \cmark &  & \cmark & \cmark & \cmark & \cmark & & Adaptive hybrid assessment and refinement \\
& EQA-RM~\citep{chen2025eqa} & Embodied QA & & \cmark & \cmark & & \cmark & \cmark & & Feedback with detailed reasoning and error critique \\
& Athena~\citep{wang2025athena} & Math Reasoning & & \cmark & & \cmark & \cmark & & & Process supervision via consistency and negative sampling \\
& VReST~\citep{zhang2025vrest} & Math Reasoning & & \cmark & & \cmark & \cmark & & & Tree-level feedback integrating utility and correctness \\
&  \citet{wang2024scaling} & Multimodal Reasoning & & \cmark & & \cmark & \cmark & & & Forward-looking rewards predicting coherence and fidelity \\
& RoVer~\citep{dai2025rover} & Vision Language Action & & \cmark & & \cmark & \cmark & & & Refines actions via verifier-guided 6D optimization \\
& VisualPRM~\citep{wang2025visualprm} & Multimodal Reasoning & & \cmark & & \cmark & \cmark & & & VisualPRM acts as a BoN verifier to improve multimodal evaluation\\
\midrule

\multirow{12}{*}{\shortstack[l]{\textbf{Iterative} \\ \textbf{Refinement}}} 
& Reflect-DiT~\citep{li2025reflect} & Image Generation & \cmark & & \cmark & & & \cmark & & Feedback-guided correction using prior outputs and text prompts \\
& \citet{zhuo2025reflection} & Image Generation & \cmark & & \cmark & & & \cmark & & Sequential reflection integrating prompts and images for correction \\
& Vidorag~\citep{wang2025vidorag} & RAG & & \cmark & & \cmark & & \cmark & & Multi-agent iterative reasoning through explore–inspect–answer cycles \\
& CyberV~\citep{meng2025cyberv} & Video Reasoning & & \cmark & & \cmark & \cmark & \cmark & & Feedback loop monitors drift and triggers adaptive self-correction \\
& GenPilot~\citep{ye2025genpilot} & Image Generation & \cmark & & \cmark & & \cmark & \cmark &  &Optimizes prompts via iterative multi-agent feedback  \\
& VideoChat-R1.5~\citep{yan2025videochat} & Video Reasoning & & \cmark & & \cmark & & \cmark & \cmark & Refines perception via iterative visual-language modeling  \\
& UI2Code$^N$~\citep{yang2025ui2code} & UI-to-Code & \cmark & & \cmark & & & \cmark & \cmark & Refines code via internalized iterative loops \\
& RAPO++~\citep{gao2025rapo++} & Video Generation & \cmark & & \cmark & & & \cmark & & Optimizes prompts via visual-semantic feedback \\
& Metal~\citep{li2025metal} & Chart Generation & \cmark & & \cmark & & & \cmark & \cmark & Refines chart code via multi-agent critique \\
& ImAgent~\citep{wang2025imagent} & Image Generation & \cmark & & \cmark & & & \cmark & & Scales generation via adaptive iterative reflection \\
& \citet{zhang2025generative} & Image Generation & \cmark & & \cmark & & & \cmark & & Refines images via verifier-driven edit prompts  \\
& DiMo-GUI~\citep{wu2025dimo} & GUI Grounding & & \cmark & & \cmark & & & \cmark & Iteratively refines coordinates through zoom-in strategy  \\
\bottomrule
\end{tabular}
}
\caption{Summary of feedback-based methods.\textbf{Gen.}: Multimodal Generation, \textbf{Reas.}: Multimodal Reasoning.} 
\label{tab:feedback_table}
\vspace{8pt}
\end{table*}

Best-of-N~\citep{cobbe2021training} employs a scoring function, or leverages an MLLM as a judge, to evaluate $N$ candidate solutions generated at inference time, selecting the highest-scoring candidate as the final output.
TTGen~\citep{qiao2025ttgen} guides the diffusion trajectory by selecting the latent variable with the highest CLIP~\citep{radford2021learning} score at each denoising step. Building on this, SANA-1.5~\citep{xie2025sana} refines candidate selection via tournament-style comparisons and Vision-Language Models (VLMs) scoring.
\citet{jang2025verifier} filter Vision-Language-Action (VLA) actions using KL divergence from a Masked Reference Distribution as the scoring metric.
Sim4Seg~\citep{song2025sim4seg} achieves cross-modal BoN by jointly scaling semantic reasoning paths and visual decoding perturbations.
To improve efficiency, CoDe~\citep{singh2025code} mitigates BoN overhead by replacing global sampling with local BoN selection every $B$ steps during reverse diffusion.
Integrating BoN with MLLM reasoning, UniGen~\citep{tian2025unigen} utilizes CoT verification to enable the model to function as both generator and verifier.

\subsubsection{Majority Voting}\label{sec:Majority Voting}
Unlike verifier-dependent BoN methods, Majority Voting~\citep{wang2022self,byun2025test} aggregates candidate solutions, selecting the most frequent or consistent output as the final prediction.
CoT-Vid~\citep{jin2025cot} substitutes traditional answer voting with character-level path clustering to ensure intermediate reasoning consistency.
For Embodied QA, EQA-RM~\citep{chen2025eqa} aggregates path evaluations via majority voting to enhance answer stability in uncertain contexts.
Video-RTS~\citep{wang2025video} achieves reliable few-shot video reasoning by combining progressive frame scaling with multi-path consistency voting.
RoboMonkey~\citep{kwok2025robomonkey} ensures stability by constructing action distributions via majority voting on Gaussian-perturbed VLA samples.

\subsection{Feedback-based Methods}\label{sec:Feedback-based Methods}

\begin{figure}[t!]
    \centering
\includegraphics[width=\columnwidth]{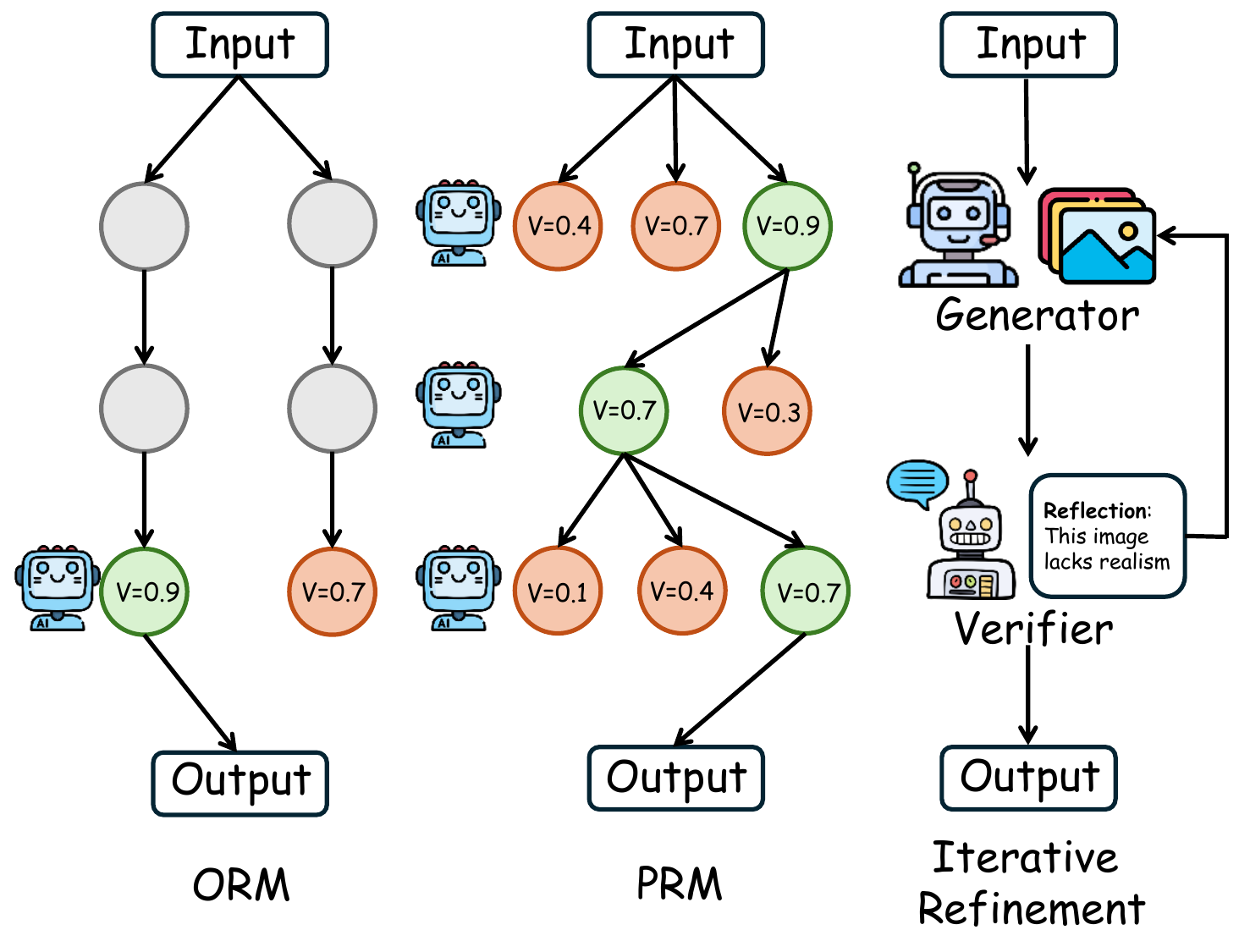}
    \caption{Illustration of feedback-based methods. ORM: Output Reward Model; PRM: Process Reward Model. Please refer to Fig.~\ref{fig:Sampling-zhixian Methods} for the legend.}
    \label{fig:Reward_zhixian Methods}
\end{figure}

Another prominent category of TTS involves feedback-based strategies, which fundamentally rely on auxiliary evaluation signals to filter, steer, or refine model outputs during inference.
Typically operating without parameter updates, these methods leverage reward models~\citep{zhao2025genprm} to guide candidate selection and intermediate reasoning, or employ iterative refinement~\citep{yao2024mulberry,zhuo2025reflection} to continuously rectify errors during generation, as illustrated in Fig.~\ref{fig:Reward_zhixian Methods}.
A comprehensive taxonomy and summary of the surveyed cases are provided in Table~\ref{tab:feedback_table}.

\subsubsection{Reward Models}\label{sec:Reward Models}

Reward models, utilizing score-based feedback to guide selection, are primarily categorized into Output Reward Models (ORMs)~\citep{xin2024advancing} and Process Reward Models (PRMs)~\citep{wang2025visualprm} based on the evaluation stage. ORMs typically evaluate final candidates generated in parallel, often coupled with strategies like BoN to identify the optimal solution.

For instance, \citet{guo2025can} employ LLaVA-OneVision as a zero-shot ORM for candidate selection, further proposing PARM/PARM++ to enable reasoning self-correction via stepwise reflection.
Similarly, \citet{chen2025eqa} introduce EQA-RM, which generates both score-based feedback and fine-grained critiques for reasoning and grounding.

In contrast, PRMs evaluate intermediate steps rather than final outcomes, offering granular feedback to facilitate exploration via strategies like Beam or Tree Search.
Specifically, Athena~\citep{wang2025athena} derives process labels from strong-weak completer consistency, enhancing PRMs through ORM initialization and negative sampling.
RoVer~\citep{dai2025rover} utilizes a plug-and-play PRM to refine 6D orientation for VLA models without retraining.
VReST~\citep{zhang2025vrest} filters reasoning paths by assessing sub-problem utility and cross-modal relevance.
Finally, \citet{wang2024scaling} uses prospective rewards to balance quality with future coherence, thereby reducing hallucination.

\subsubsection{Iterative Refinement}\label{sec:Iterative Refinement}
Distinct from reward modeling, iterative refinement emphasizes an explicit "generate-evaluate-correct" inference loop to progressively optimize outputs.
Reflect-DiT~\citep{li2025reflect} integrates VLM feedback with past outputs to guide the Diffusion Transformer in iteratively refining image generation.
For UI tasks, UI2Code$^N$~\citep{yang2025ui2code} employs an internalized "generate-observe-correct" visual feedback mechanism to achieve iterative code refinement.
Targeting video attention drift, CyberV~\citep{meng2025cyberv} employs a sensor-controller feedback loop for dynamic rectification.

Furthermore, Metal~\citep{li2025metal} and Vidorag~\citep{wang2025vidorag} extend iterative refinement to multi-agent frameworks, utilizing collaborative feedback to optimize chart generation and visual document reasoning. 
Meanwhile, VideoChat-R1.5~\citep{yan2025videochat} and DiMo-GUI~\citep{wu2025dimo} employ iterative perception and dynamic zooming to refine key regions, excelling in video spatiotemporal modeling and GUI grounding, respectively.
Similarly, to transcend one-pass generation limits, RAPO++ \cite{gao2025rapo++}, \citet{zhang2025generative}, GenPilot~\citep{ye2025genpilot}, and ImAgent~\citep{wang2025imagent} employ iterative prompt optimization, dynamically rewriting inputs via visual verification to progressively enhance quality.

\subsection{Search-based Methods}\label{sec:Search-based Methods}

\begin{table*}[t!]
\centering
\resizebox{\textwidth}{!}{
\begin{tabular}{ll l cc ccc l}
\toprule
\multirow{2}{*}{\textbf{Category}} & \multirow{2}{*}{\textbf{Method}} & \multirow{2}{*}{\textbf{Task}} & \multicolumn{2}{c}{\textbf{Domain}} & \multicolumn{3}{c}{\textbf{Search Strategy}} & \multirow{2}{*}{\textbf{Approach Description}} \\
\cmidrule(lr){4-5} \cmidrule(lr){6-8}
 & & & \textbf{Gen.} & \textbf{Reas.} & \textbf{Pruning} & \textbf{Backtrack} & \textbf{Dynamic} & \\
\midrule

\multirow{5}{*}{\textbf{Beam Search}} 
& \citet{cong2025can} & Video Generation & \cmark & & \cmark & & & Explores trajectories via combined Top-K sampling and beam search \\
& \citet{oshima2025inference} & Video Generation & \cmark & & \cmark & & & Selects diffusion trajectories using foresight-guided beam search \\
&  \citet{li2025dynamic} & Generation Alignment & \cmark & & \cmark & & \cmark & Dynamically schedules tree expansion with novel heuristics \\
& LLaVA-CoT~\citep{xu2025llava} & Multimodal Reasoning & & \cmark & \cmark & \cmark & & Generates candidates at each stage and retraces if needed \\
& MindJourney~\citep{yang2025mindjourney} & Spatial Reasoning & & \cmark & \cmark & & & Uses world model to guide beam search in spatial reasoning \\
\midrule
\multirow{7}{*}{\textbf{Tree Search}} 
& VReST~\citep{zhang2025vrest} & Math Reasoning & & \cmark & & \cmark &  & Combines MCTS with self-rewards to enhance reasoning \\
& Visuothink~\citep{wang2025visuothink} & Multimodal Reasoning & & \cmark & & \cmark &  & Uses multimodal tree search with rollback for visual-text reasoning \\
& \citet{dong2024progressive} & Math Reasoning & & \cmark & & \cmark &  & Integrates active retrieval into MCTS for dynamic knowledge \\
& Mulberry~\citep{yao2024mulberry} & Multimodal Reasoning & & \cmark & \cmark & \cmark & \cmark & Incorporates collective learning into MCTS for efficient reasoning \\
& ZoomEye~\citep{shen2025zoomeye} & Image Reasoning & & \cmark & & \cmark & \cmark & Refines perception via hierarchical tree search  \\
& VLA-Reasoner~\citep{guo2025vla} & Vision Language Action & & \cmark & & \cmark & & Optimizes actions via MCTS with world models  \\
& AKEYS~\citep{fan2025agentic} & Video Reasoning & & \cmark & & \cmark & \cmark & Searches keyframes via agent-guided tree search  \\
\midrule
\multirow{8}{*}{\shortstack{\textbf{Heuristic and} \\ \textbf{Adaptive Search}}} 
& \citet{jajal2025inference} & Generation Alignment & \cmark & & \cmark & &  & Introduces evolutionary search for gradient-independent alignment \\
& \citet{he2025scaling} & Multimodal Generation & \cmark & & \cmark & &  & Uses denoising selection and mutation mechanisms \\
&\citet{ramesh2025test} & Image Generation & \cmark & & \cmark & &  & Models denoising as a multi-armed bandit problem via $\varepsilon$-greedy search \\
& \citet{lee2025adaptive} & Multimodal Reasoning & & \cmark & \cmark & & \cmark & Uses adaptive cyclic diffusion for dynamic resource allocation \\
& Video-RTS~\citep{wang2025video} & Video Reasoning & & \cmark & & & \cmark & Adaptively adds video frames based on output consistency \\
& VideoICL~\citep{kim2025videoicl} & Video Reasoning & & \cmark & & & \cmark & Adapts context via relevant example retrieval  \\
& TTS-VAR~\citep{chen2025tts} & Image Generation & \cmark & & \cmark & & \cmark & Scales generation via adaptive batch strategies  \\
& ImagerySearch~\citep{wu2025imagerysearch} & Video Generation & \cmark & & & & \cmark & Aligns generation via dynamic search spaces  \\
\bottomrule
\end{tabular}
}
\caption{Summary of search-based methods. \textbf{Gen.}: Multimodal Generation, \textbf{Reas.}: Multimodal Reasoning. \textbf{Search Strategy}: \textbf{Pruning} filters parallel candidates; \textbf{Backtrack} enables recursive state rollback; \textbf{Dynamic} dynamically adjusts compute budget (e.g., depth/width) based on difficulty.}
\label{tab:search_table}
\vspace{6pt}
\end{table*}

\begin{figure}[t!]
    \centering
\includegraphics[width=\columnwidth]{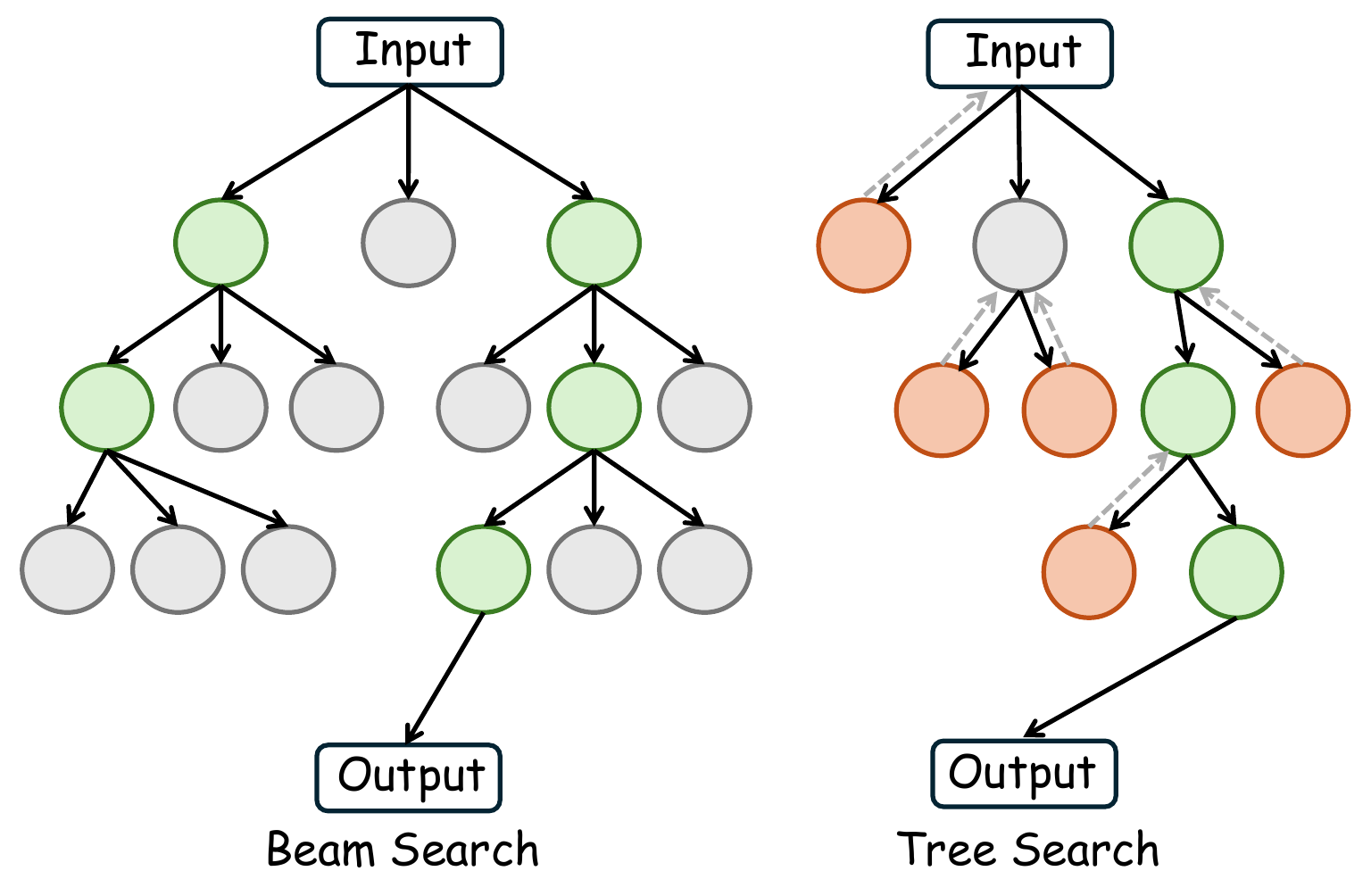}
    \caption{Illustration of search-based methods. Please refer to Fig.~\ref{fig:Sampling-zhixian Methods} for the legend.}
    \label{fig:Search-zhixian Methods}
\end{figure}

Search-based TTS enables "planning-based exploration" through systematic or heuristic trajectory exploration, leveraging structured search mechanisms during inference rather than relying solely on stochastic sampling or post-hoc refinement.  
As illustrated in Fig.~\ref{fig:Search-zhixian Methods}, we categorize search-based TTS approaches into three primary streams: Beam Search, Tree Search, and Heuristic and Adaptive Search.
A comprehensive classification and summary of all surveyed works is detailed in Table~\ref{tab:search_table}.

\subsubsection{Beam Search}\label{sec:Beam Search}
Beam Search~\citep{welleck2022naturalprover} maintains multiple candidate paths during inference, pruning low-scoring candidates at each generation step to balance computational efficiency with search breadth.
\citet{cong2025can} integrate Top-K sampling with beam search for sequence exploration. Similarly, \citet{oshima2025inference} extend this logic to diffusion models via Diffusion Latent Beam Search, utilizing lookahead estimators to optimize latent trajectories based on alignment rewards.
\citet{li2025dynamic} optimizes efficiency by dynamically adjusts tree and beam widths based on noise levels.
Focusing on trajectory quality, LLaVA-CoT~\citep{xu2025llava} integrates beam search with backtracking to regenerate from prior stages when local candidates underperform. 
MindJourney~\citep{yang2025mindjourney} integrates a world model to simulate future views for each beam candidate, employing VLM valuation to dynamically plan optimal spatial paths.

\subsubsection{Tree Search}\label{sec:Tree Search}

Tree Search systematically explores the solution space by recursively generating, branching, and backtracking through candidate nodes during inference~\citep{luo2024improve}.
For hierarchical exploration, AKEYS~\citep{fan2025agentic} guides keyframe refinement via agent-driven binary search, while ZoomEye~\citep{shen2025zoomeye} implements multi-scale tree search with lookahead and backtracking mechanisms for fine-grained perception.
VisuoThink~\citep{wang2025visuothink} abstracts "slow thinking" into an interleaved visual-textual tree search, employing predictive rollback to simulate and prioritize promising branches.
Along this line, recent works leverage Monte Carlo Tree Search (MCTS) to enable effective multimodal test-time scaling. While VReST~\citep{zhang2025vrest} and VLA-Reasoner~\citep{guo2025vla} drive scaling by leveraging internal self-rewards and world model simulations to guide the MCTS process, \citet{dong2024progressive} and Mulberry~\citep{yao2024mulberry} enhance inference robustness by incorporating external retrieval and collective multi-model learning into the MCTS framework.

\subsubsection{Heuristic and Adaptive Search}\label{sec:Heuristic and Adaptive Search}
Distinct from beam or tree search, heuristic and adaptive search leverage dynamic mechanisms to efficiently navigate the solution space without fixed structural constraints. 
Within this paradigm, recent works exploit evolutionary algorithms for test-time scaling. Specifically, \citet{jajal2025inference} conduct black-box evolutionary search over latent spaces to enable gradient-free alignment, whereas \citet{he2025scaling} utilizes denoising-based selection and mutation mechanisms for exploration.
To optimize efficiency, recent works dynamically allocate computational budgets based on instance difficulty. For instance, VideoICL~\citep{kim2025videoicl} and \citet{lee2025adaptive} implement adaptive termination mechanisms, dynamically halting the search process once prediction confidence or consistency criteria are met. Similarly, Video-RTS~\citep{wang2025video} adopts a sparse-to-dense strategy, iteratively incorporating frames guided by output consistency to tailor computation for specific video queries.
Beyond efficiency, adaptive search modulates the exploration–exploitation trade-off to enhance generative alignment. \citet{ramesh2025test} and ImagerySearch~\citep{wu2025imagerysearch} leverage contextual bandits or mental imagery simulation to dynamically adjust the search space, balancing global exploration with local refinement. Similarly, TTS-VAR~\citep{chen2025tts} orchestrates a coarse-to-fine search trajectory, adaptively shifting from early diversity-oriented clustering to late-stage reward-guided resampling.

\subsection{Comparative Analysis and Trade-offs}

The effectiveness of multimodal TTS methods is closely related to task characteristics and modality structure. For multimodal generation tasks, sampling-based methods and iterative refinement are often more suitable than exhaustive search, because visual generation is mainly judged by final output quality rather than explicitly verifiable intermediate states. By contrast, search-based methods are more advantageous for multimodal reasoning tasks, such as mathematical and spatial reasoning, where intermediate reasoning steps are more structured and partially verifiable. This allows the model to prune erroneous branches and backtrack when necessary. Feedback-based methods lie between these two extremes, providing more targeted guidance than pure sampling while remaining less expensive than full search.

These categories also differ in trade-offs between performance and efficiency.
Sampling-based methods are easy to parallelize, but their gains often diminish as the number of candidates increases. Feedback-based methods can improve alignment and reliability more directly, but they introduce sequential latency that depends heavily on the verifier or judge model. Search-based methods incur the highest computational overhead due to repeated branching, evaluation, and rollback, yet they are often the most effective when accuracy is prioritized and process supervision is available, especially for long-chain multimodal reasoning.

\vspace{5pt}
\section{Applications}\label{sec:Applications and Benchmarks}
This section highlights representative applications of TTS for MFMs within Multimodal Generation and Multimodal Reasoning, analyzing their domain-specific scaling strategies. 
For a comprehensive review of benchmarks, please refer to Appendix~\ref{sec:Benchmarks} and the detailed summary in Appendix Table~\ref{tab:benchmark-summary}.
\subsection{Multimodal Generation}
\paragraph{Image Generation}
TTS in image generation primarily exploits the trade-off between inference compute and visual-semantic alignment through search and refinement methods. 
The dominant BoN strategy allocates computational budget to parallel sampling, utilizing ORMs or scoring functions~\citep{qiao2025ttgen,tian2025unigen} to filter the optimal candidate from a vast generation space.
Complementary to sampling, iterative refinement methods dynamically optimize input prompts or conditioning signals based on self-correction feedback~\citep{li2025reflect,wang2025imagent}.

\paragraph{Video Generation}
To ensure temporal consistency and motion smoothness, TTS strategies scale inference compute using search-based methods. 
Approaches such as Beam Search and Tree Search explore spatiotemporal trajectories, utilizing PRM to score and select optimal frame sequences~\citep{cong2025can,liu2025video}. 
This mechanism allows for pruning incoherent paths to mitigate error propagation and maintain semantic stability across long durations.

\subsection{Multimodal Reasoning}
\paragraph{Video Reasoning}
To tackle long-context reasoning, TTS strategies primarily rely on search-based methods or iterative refinement. 
Search-based approaches actively navigate the temporal space to retrieve segments most relevant to the query~\citep{fan2025agentic}. 
Alternatively, iterative refinement strategies focus on extracting key visual evidence and explicitly judging its relevance to the inquiry~\citep{yan2025videochat}, thereby concentrating computational resources on critical events while filtering out redundant information.

\paragraph{Vision Language Action}
To enhance precision in physical control and long-horizon planning, TTS strategies in VLA primarily adopt sampling-based paradigms and tree search algorithms. 
Sampling-based methods leverage parallel computation to generate multiple candidate action trajectories, selecting the most robust execution path via consensus or scoring mechanisms~\citep{jang2025verifier,kwok2025robomonkey}. 
Furthermore, tree search approaches are applied to systematically explore the expansive action solution space, enabling models to perform lookahead planning and optimize multi-step decisions against complex environmental constraints~\citep{guo2025vla}.

\paragraph{Math Reasoning}
TTS for multimodal math reasoning predominantly relies on MCTS~\citep{zhang2025vrest,wang2025visuothink}. 
By performing look-ahead simulations, MCTS systematically explores reasoning trajectories to optimize solution paths. 
Crucially, it incorporates prediction roll-back mechanisms to revert invalid steps, enabling the model to recover from intermediate visual misinterpretations or calculation errors and re-navigate toward correct solutions.

\vspace{9pt}

\section{Challenges and Future Directions}\label{sec:Challenges and Future Directions}

\vspace{3pt}

\paragraph{Hybrid Scaling}
Current multimodal TTS approaches typically rely on singular strategies, thereby underutilizing the synergistic potential of complementary mechanisms.
Moderate increases in sampling paths enhance performance but incur significant costs at scale; conversely, relying solely on search strategies compromises efficiency by requiring optimal solution discovery within vast search spaces.

Hybrid scaling strategies have demonstrated potential in LLMs.
For instance, Marco-o1~\citep{zhao2024marco} integrates MCTS with reflection mechanisms to dynamically plan inference paths via confidence-based search.
However, systematic exploration of such hybrid TTS strategies within multimodal contexts remains nascent.
Consequently, future research should explore hybrid TTS frameworks integrating multiple scaling mechanisms to balance performance and efficiency.

\paragraph{Error Propagation}
In multimodal reasoning, particularly during long-chain or cross-frame video tasks, early missteps can trigger an \emph{Error Snowballing Effect}~\citep{gan2025rethinking}.
Once initial visual or semantic errors occur, subsequent reasoning steps often amplify these deviations, precipitating catastrophic failure.
Current multimodal TTS research lacks systematic mechanisms to arrest error propagation in long-chain reasoning, as most approaches remain confined to output-level optimization.
A promising solution is employing trajectory-correcting reward models to detect and fix deviations during inference. Furthermore, establishing critical node verification mechanisms is also expected to ensure consistency across long-chain multimodal reasoning.

\paragraph{Hallucination Control}
Current MLLMs frequently hallucinate object attributes or relationships, decoupling generations from perceptual reality~\citep{huang2024opera}.
Current approaches largely rely on output-level post-hoc checks, such as detecting factual consistency, to address hallucinations.
However, such retrospective correction fails to fundamentally constrain the formation and propagation of hallucinations during inference.
Future research should pivot to process-level suppression and dynamic verification. Specifically, cross-modal consistency strategies can mitigate modality misalignment via reciprocal visual-textual checks. 
Furthermore, multi-level alignment is crucial to fuse perceptual and semantic constraints, ensuring both visual fidelity and logical precision throughout the inference trajectory.

\vspace{8pt}
\section{Conclusion}\label{sec:Conclusions}

This paper presents the first systematic review of TTS for MFMs. Grounded in fundamental principles, we establish a unified taxonomy that categorizes existing methodologies into sampling-based, feedback-based, and search-based strategies.We also formalize TTS in MFMs, distinguishing it from memory and adaptation methods, and highlight its unique multimodal challenges compared to TTS in LLMs.
Furthermore, we scrutinize application patterns across multimodal generation and reasoning tasks and collate relevant benchmarks to serve as a comprehensive reference for future inquiry.
Finally, we identify key challenges and outline promising avenues for future research, including hybrid scaling strategies, hallucination mitigation, and addressing error accumulation.
Pursuing these avenues will enable efficient, robust MFMs, ultimately paving the way for AGI with advanced cognitive reasoning.


\section{Limitations}
While this survey strives to offer a comprehensive overview of Test-time Scaling in Multimodal Foundation Models, we acknowledge several limitations.
First, regarding the scope, despite the broad connotation of the term \emph{multimodal}, this work exclusively focuses on vision-language modalities (i.e., images and videos) and does not cover audio or other sensory inputs.
Second, to prioritize domain-specific strategies, we do not conduct a comparative analysis with scaling techniques designed specifically for pure LLMs.
Finally, given the rapid evolution of this field, some of the most recent advancements may inevitably be omitted, and due to space constraints, we cannot claim an exhaustive coverage of every existing technique or application.

\section*{Acknowledgments}
This work was supported in part by National Natural Science Foundation of China under Grant No. 62272494, and Guangdong Basic and Applied Basic Research Foundation under Grant No. 2023A1515012845.

\bibliography{custom.bib}

\newpage

\numberwithin{equation}{section}
\numberwithin{table}{section}
\numberwithin{figure}{section}

\renewcommand{\theequation}{\thesection.\arabic{equation}}
\renewcommand{\thetable}{\thesection.\arabic{table}}
\renewcommand{\thefigure}{\thesection.\arabic{figure}}

\appendix

\section*{Appendix}

\section{Benchmarks}
\label{sec:Benchmarks}
Complementing the discussion on application-specific scaling strategies in the main text, this appendix provides a detailed overview of the benchmarks used to evaluate TTS performance.
We categorize these benchmarks into Multimodal Generation and Multimodal Reasoning.
Table~\ref{tab:benchmark-summary} presents a comprehensive summary of these datasets, highlighting their specific tasks relevant to TTS deployment.

\subsection{Multimodal Generation}
\paragraph{Image Generation}
MSCOCO~\citep{lin2014microsoft} serves as a seminal benchmark for quantifying the semantic consistency and perceptual quality of generated images.
To address MSCOCO's limited prompt diversity, DrawBench~\citep{saharia2022photorealistic} and GenEval~\citep{ghosh2023geneval} were introduced for fine-grained object- and attribute-level evaluation.
T2I-CompBench~\citep{huang2023t2i} explicitly targets compositional capabilities under varying attribute and spatial relationship constraints.
Meanwhile, DPGBench~\citep{hu2024ella} addresses the lack of high-density prompt analysis by systematically benchmarking model performance on complex inputs.

\paragraph{Video Generation}
VBench~\citep{huang2024vbench} aims to establish a comprehensive suite anchored in human preference validation.
Advancing this, VBench2~\citep{zheng2025vbench} and MovieGenVideoBench~\citep{polyak2024movie} incorporate "intrinsic reality" to assess adherence to physical laws.
In parallel, VideoGen-Eval~\citep{yang2025videogen} introduces an agent-based dynamic framework to better cover out-of-distribution scenarios and complex prompts.
Targeting long-form generation, MovieBench~\citep{wu2025moviebench} evaluates multi-scene narratives, cross-character consistency, and hierarchical structures.

\begin{table*}[!t]
\begin{adjustbox}{width=\textwidth,keepaspectratio}
\begin{tabular}{cccccc}
\toprule
\textbf{Type} & \textbf{Benchmark} & \textbf{Size} & \textbf{Metrics} & \textbf{Features} \\
\midrule
\multicolumn{5}{c}{\textbf{Multimodal Generation}} \\
\midrule
\multirow{5}{*}{Image Generation} & MSCOCO~\citep{lin2014microsoft} & 2.5M & CLIPScore & Semantic consistency \\
& DrawBench~\citep{saharia2022photorealistic} & 200 & Human alignment & Fine-grained evaluation \\
& GenEval~\citep{ghosh2023geneval} & 6K & Human alignment & Fine-grained evaluation \\
& T2I-CompBench~\citep{huang2023t2i} & 6K & BLIP-VQA & Compositionality evaluation\\
& DPGBench~\citep{hu2024ella} & 1K & PLUG score & Dense-Prompt evaluation \\

\midrule
\multirow{5}{*}{Video Generation} & VBench~\citep{huang2024vbench} & Varied & Task-specific scores & Fine-Grained evaluation \\
& VBench2~\citep{zheng2025vbench} & Varied & Task-specific scores & Intrinsic faithfulness evaluation \\
& MovieGenVideoBench~\citep{polyak2024movie} & 1003 & Task-specific scores & Fine-Grained evaluation \\
& VideoGen-Eval~\citep{yang2025videogen} & 12000 & Task-specific scores & Agent-based evaluation \\
& MovieBench~\citep{wu2025moviebench} & 91 & Task-specific scores & Long-video evaluation \\

\midrule
\multicolumn{5}{c}{\textbf{Multimodal Reasoning}} \\
\midrule
\multirow{4}{*}{Spatial Reasoning} & SAT~\citep{ray2024sat} & 175K & Task-specific scores & Dynamic spatial reasoning \\
& SpatialRGBT~\citep{cheng2024spatialrgpt} & 1406 & Task-specific scores & 3D-Spatial cognition evaluation \\
& VSI-Bench~\citep{yang2025thinking} & 5K & Accuracy\&MRA & Video spatial reasoning \\
& SPAR-Bench~\citep{zhang2025flatland} & 7207 & Accuracy\&MRA & Multi-view spatial reasoning \\

\midrule
\multirow{3}{*}{GUI Grounding} & ScreenSpot~\citep{cheng2024seeclick} & 1.2K & Accuracy & Various GUI platforms \\
& ScreenSpot-V2~\citep{wu2024atlas} & 1.2K & Accuracy & Correcting annotation errors\\
& ScreenSpot-Pro~\citep{li2025screenspot} & 1581 & Accuracy & High-resolution GUI evaluation \\

\midrule
\multirow{5}{*}{Math Reasoning} & MathVista~\citep{lu2023mathvista} & 6141 & Accuracy & Visual-math integration \\
& MathVision~\citep{wang2024measuring} & 3040 & Accuracy & Multimodal math reasoning \\
& MathVerse~\citep{zhang2024mathverse} & 15K & Accuracy & Cross-Version evaluation \\
& WeMath~\citep{qiao2024we} & 6.5K & Score & Stepwise reasoning analysis \\
& DynaMath~\citep{zou2024dynamath} & 5K & Accuracy & Multimodal math reasoning \\

\midrule
\multirow{7}{*}{Video Reasoning} & MVBench~\citep{li2024mvbench} & 4K & Accuracy & Automated QA evaluation \\
& CG-Bench~\citep{chen2024cg} & 12129 & Accuracy\&IoU & Long-video understanding \\
& MMBench-Video~\citep{fang2024mmbench} & 600 & GPT-4 score & Long-video understanding \\
& LongVideoBench~\citep{wu2024longvideobench} & 6678 & Accuracy & Long-video understanding \\
& MLVU~\citep{zhou2025mlvu} & 3102 & Accuracy & Multi-type video evaluation \\
& Video-MME~\citep{fu2025video} & 2700 & Accuracy & Expert-annotated evaluation\\
& Video-MMMU~\citep{hu2025video} & 900 & Accuracy & Multi-Discipline video understanding \\

\midrule
\multirow{3}{*}{Medical Reasoning} & OmniMedVQA~\citep{hu2024omnimedvqa} & 127K & QA\&Prefix-based score & Comprehensive medical VQA \\
& GMAI-MMBench~\citep{ye2024gmai} & 26K & Accuracy & Comprehensive medical reasoning \\
& MedXpertQA~\citep{zuo2025medxpertqa} & 4460 & Accuracy & Expert-level medical reasoning \\

\midrule
\multirow{3}{*}{Vision Language Action} 
& SimplerEnv~\citep{li2024evaluating} & 1500 & MMRV & Visual-Spatial Robustness \\
& LIBERO~\citep{liu2023libero} & 130 & FWT\&AUC & Multitask Knowledge Transfer \\
& CALVIN~\citep{mees2022calvin} & 20K & Success Rate & Long-Horizon Generalization \\
& ManiSkill2~\citep{gu2023maniskill2} & 4M & Success Rate & Generalizable Dynamic Manipulation \\

\bottomrule
\end{tabular}
\end{adjustbox}
\caption{Taxonomy of benchmarks for evaluating multimodal generation and reasoning capabilities. } 
\label{tab:benchmark-summary}
\end{table*}

\begin{table*}[!t]
\centering
\small
\renewcommand{\arraystretch}{1.2}
\begin{tabular}{p{2.6cm}p{2.4cm}p{3.5cm}p{3.2cm}}
\toprule
\textbf{Category} & \textbf{Test-Time Scaling} & \textbf{Test-Time Memory} & \textbf{TTT/Adaptation} \\
\midrule
Scaled dimension & Compute & Dynamic context / cache & Parameters \\
Typical mechanisms & Sampling, Search & KV cache, RAG & fine-tuning, LoRA \\
Metrics & FLOPs, latency & Context length, cache size & Gradient steps \\
\bottomrule
\end{tabular}
\caption{A compact comparison of test-time scaling, test-time memory, and test-time training(TTT)/adaptation.} \vspace{-2pt}
\label{tab:tts_scope}
\end{table*}

\subsection{Multimodal Reasoning}
\paragraph{Spatial Reasoning}
SAT~\citep{ray2024sat} introduces a dataset spanning static and dynamic scenes, emphasizing the role of motion and temporal dynamics in spatial understanding.
Subsequently, SpatialRGBT~\citep{cheng2024spatialrgpt} advances cross-scene spatial evaluation by integrating ground-truth 3D annotations across indoor, outdoor, and simulated environments.
VSI-Bench~\citep{yang2025thinking} further proposed a video-based framework to evaluate spatial intelligence across configuration, measurement, and spatiotemporal tasks.
Finally, SPAR-Bench~\citep{zhang2025flatland} expands task scope and supports single/multi-view inputs, overcoming prior limitations in multi-view and local structure modeling.

\paragraph{GUI Grounding}
ScreenSpot~\citep{cheng2024seeclick} benchmarks cross-platform grounding, while ScreenSpot-V2~\citep{wu2024atlas} enhances reliability by rectifying annotation errors. Furthermore, ScreenSpot-Pro~\citep{li2025screenspot} extends to high-resolution professional scenarios across diverse OSs, aligning evaluation with realistic, complex tasks.

\paragraph{Math Reasoning}
MathVista~\citep{lu2023mathvista} pioneered the systematic integration of diverse multimodal math tasks but remains predominantly static. MathVision~\citep{wang2024measuring} further enhances complexity and diversity by incorporating multidisciplinary competition problems.
Building on this, MathVerse~\citep{zhang2024mathverse} scrutinizes the model's genuine reliance on visual cues by constructing multi-version problem variations.
Diverging from outcome-oriented evaluations, WeMath~\citep{qiao2024we} adopts a knowledge-centric approach to elucidate intrinsic reasoning mechanisms. DynaMath~\citep{zou2024dynamath} addresses prior gaps in robustness and generalization by employing a dynamic generation mechanism.

\paragraph{Video Reasoning}
Early benchmarks, such as MVBench~\citep{li2024mvbench}, primarily focused on bridging the gap between static imagery and dynamic video.
Subsequent research pivoted to long-video scenarios, emphasizing long-temporal and multi-shot reasoning, represented by MMBench-Video~\citep{fang2024mmbench}, LongVideoBench~\citep{wu2024longvideobench}, and CG-Bench~\citep{chen2024cg}.
More recent efforts have further expanded benchmark scope and task diversity, exemplified by MLVU~\citep{zhou2025mlvu}, Video-MME~\citep{fu2025video}, and Video-MMMU~\citep{hu2025video}.

\paragraph{Medical Reasoning}
Addressing data scarcity, OmniMedVQA~\citep{hu2024omnimedvqa} pioneered the integration of diverse medical imagery with VQA, establishing a comprehensive evaluation framework across multiple anatomical regions.
Built on this, GMAI-MMBench~\citep{ye2024gmai} introduces multi-granularity perception and hierarchical task taxonomies, bridging prior gaps in clinical relevance and evaluation dimensionality.
MedXpertQA~\citep{zuo2025medxpertqa} further incorporates licensure-level QA and authentic clinical contexts to simulate expert-grade diagnostic reasoning.

\paragraph{Vision Language Action}
Addressing sequential instruction following, CALVIN~\citep{mees2022calvin} benchmarks long-horizon language-conditioned tasks, utilizing the ABC$\to$D protocol to evaluate zero-shot generalization across scenes.
Targeting continuous adaptation, LIBERO~\citep{liu2023libero} introduces a framework for lifelong robot learning, focusing on knowledge transfer across diverse multitask scenarios ranging from spatial reasoning to long-horizon manipulation.
Furthermore, ManiSkill2~\citep{gu2023maniskill2} targets generalized manipulation capabilities, utilizing fully dynamic simulations and diverse object variations to evaluate policy robustness across extensive task configurations.
Finally, SimplerEnv~\citep{li2024evaluating} mitigates physical testing limitations by developing a high-fidelity Real-to-Sim platform, serving as a scalable and reproducible proxy for evaluating generalist policies.

\end{document}